\documentclass[journal]{IEEEtran}

\usepackage[linesnumbered,lined,noend,ruled]{algorithm2e}
\usepackage{algorithmic}
\usepackage{amsmath,amssymb}
\usepackage{subfigure}
\usepackage{amstext}
\usepackage{amsfonts}
\usepackage{bbm}
\usepackage{color}
\usepackage{amssymb}
\usepackage{float}

\usepackage{color}
\definecolor{red}{RGB}{255,0,0}

\usepackage{multirow}
\usepackage[export]{adjustbox}
\usepackage{ragged2e}
\usepackage{array}

\usepackage{url}

\usepackage{xspace}

\newcommand{\ie}{\textit{i.e.}\xspace}
\newcommand{\eg}{\textit{e.g.}\xspace}
\newcommand{\etal}{\textit{et al.}\xspace}
\newcommand{\etc}{\textit{etc.}\xspace}

\begin{document}
\title{Arbitrary Shape Text Detection via Boundary Transformer}

\author{Shi-Xue Zhang, Chun Yang, Xiaobin Zhu, Xu-Cheng Yin,~\IEEEmembership{Senior Member,~IEEE}

\thanks{Corresponding authors: Xiaobin Zhu.}
\thanks{S. Zhang, X. Zhu, and C. Yang are with the School of Computer and Communication Engineering, University of Science and Technology Beijing (USTB), Beijing, 100083, China (e-mail: zhangshixue111@163.com; chunyang@ustb.edu.cn; zhuxiaobin@ustb.edu.cn.)}
\thanks{X. Yin  is with the School of Computer and Communication Engineering, and Institute of Artificial Intelligence, University of Science and Technology Beijing (USTB), also with USTB-EEasyTech Joint Lab of Artificial Intelligence, Beijing, 100083, China (e-mail: xuchengyin@ustb.edu.cn).}
\thanks{Manuscript received *** **, 2022; revised *** **, 2022.}}

\markboth{Journal of \LaTeX\ Class Files,~Vol.~14, No.~8, August~2023}%
{Shell \MakeLowercase{\textit{et al.}}: Bare Demo of IEEEtran.cls for IEEE Journals}

\maketitle

\begin{abstract}
  In arbitrary shape text detection, locating accurate text boundaries is challenging and non-trivial.  
  Existing methods often suffer from indirect text boundary modeling or complex post-processing. 
  In this paper, we systematically present a unified coarse-to-fine framework via boundary learning for arbitrary shape text detection, which can accurately and efficiently locate text boundaries without post-processing.
  In our method, we explicitly model the text boundary via an innovative iterative boundary transformer in a coarse-to-fine manner. In this way, our method can 
  directly gain accurate text boundaries and abandon complex post-processing to improve efficiency. Specifically, our method mainly consists of a feature extraction backbone, a boundary proposal module, and an iteratively optimized boundary transformer module. The boundary proposal module consisting of multi-layer dilated convolutions will predict important prior information (including classification map, distance field, and direction field) for generating coarse boundary proposals while guiding the boundary transformer's optimization. The boundary transformer module adopts an encoder-decoder structure, in which the encoder is constructed by multi-layer transformer blocks with residual connection while the decoder is a simple multi-layer perceptron network (MLP). Under the guidance of prior information, the boundary transformer module will gradually refine the coarse boundary proposals via iterative boundary deformation. Furthermore, we propose a novel boundary energy loss (BEL) that introduces an energy minimization constraint and an energy monotonically decreasing constraint to further optimize and stabilize the learning of boundary refinement. Extensive experiments on publicly available and challenging datasets demonstrate the state-of-the-art performance and promising efficiency of our method. The code and model are available at: https://github.com/GXYM/TextBPN-Puls-Plus.
\end{abstract}

\begin{IEEEkeywords} Arbitrary shape text detection, boundary proposal, boundary transformer, boundary energy.
\end{IEEEkeywords}

\IEEEpeerreviewmaketitle


\section{Introduction}\label{sec:introduction}
\IEEEPARstart{S}{cene} text detection is an essential and fundamental task in
computer vision as it is an important step in various text-related applications, such as text recognition, text retrieval, text visual question answering, and online education. Benefiting from the rapid development of CNN-based object detection and instance segmentation, scene text detection has witnessed significant progress and achieved impressive performance for texts of regular shape or aspect ratio. As one of the most challenging tasks in text detection, arbitrary shape text detection has received ever-increasing interest in both research and industrial communities.

Distinguishing from general object detection with bounding boxes, arbitrary shape text detection should explore irregular boundaries for each individual text. Connected Component (CC) based methods~\cite{TextSnake, DRRG} model text instances with sequential components or local boxes. Segmentation-based methods~\cite{PSENet_v2, DB} model arbitrary shape text instances at pixel-level mask predictions and detect the text boundaries with the edge of masks. Both the CC-based and segmentation-based methods model text instances from local perspectives (local components or pixels) instead of modeling text boundaries directly. Consequently, they tend to neglect the global geometric distribution of the overall layout of the text boundary, causing two main problems: one problem is that they are sensitive to noise due to the homogenous texture within text regions; the other problem is that they would rely on complicated and heuristic post-processings for generating accurate text boundaries.

\begin{figure}[tbp]
	\subfigcapskip=-5pt
	\centering
	\includegraphics[width=1.0\linewidth]{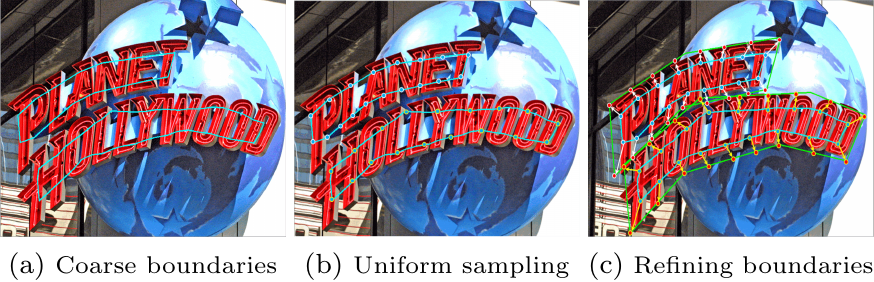}
	\centering
	\caption{ Illustration of refining text boundaries via boundary transformer. (a) Generating coarse text boundaries; (b) Uniform sampling on boundaries; (c) Refining coarse text boundaries. The green contours are annotations.}
	\label{fig:introduction}
	\vspace{-1.6em}
\end{figure}

Recently, a multiplicity of contour-based solutions~\cite{CVPR19_ATRR,ContourNet, ABCNet, FCENet, TextRay, PCR}
have been proposed to detect the boundaries of arbitrary shape texts directly and achieve promising performance. ABCNet~\cite{ABCNet} and FCENet~\cite{FCENet} respectively model text instance contours with Bezier-Curve and Fourier-Curve for effectively regressing closed contours via a progressive approximation strategy. TextRay~\cite{TextRay} proposes a single-shot framework in a polar system to predict geometric parameters and output simple polygon detections at one pass with the NMS post-processing step. Some other methods~\cite{CVPR19_ATRR,ContourNet} adopt top-down detection frameworks to regress the key points on text contours by the  ROI operation.
As claimed in~\cite{PCR}, these methods only perceive scene texts with complex geometry layouts at one stage, resulting in inaccurate localization. However, this is inconsistent with the human visual system in which looking more than once is usually required~\cite{CVPR19_LOMO}. Therefore, a coarse-to-fine framework may be more reasonable for arbitrary shape text detection. PCR~\cite{PCR} designs a coarse-to-fine pipeline with a progressive contour regression strategy based on a top-down detection framework. Specifically, it first regresses the global contours of the horizontal proposals to the corner points of oriented proposals. Then the contours of oriented proposals are iteratively regressed to arbitrarily shaped ones by boundary deformation.  Although PCR~\cite{PCR} achieves promising performance, the efficiency of top-down structure always suffers from the too complex pipeline and time-consuming NMS process, and the boundary deformation network seriously affects the accuracy of boundary detection.

In this paper, we systematically present a unified coarse-to-fine framework via boundary learning for arbitrary shape text detection, which can accurately and efficiently locate text boundaries without post-processing. In our method, we explicitly model the text boundary via an innovative iterative boundary transformer in a coarse-to-fine manner. In this way, our method can directly generate accurate text boundaries and abandon complex post-processing to improve efficiency. Specifically, our method mainly consists of a feature extraction backbone, a boundary proposal module, and an iteratively optimized boundary transformer module. The boundary proposal module consisting of multi-layer dilated convolutions will compute important prior information (including classification map, distance field, and direction field) for generating coarse boundary proposals while guiding the boundary transformer's optimization. As shown in Fig.~\ref{fig:introduction} (a) and (c), the coarse boundary proposals can roughly locate texts and separate neighboring texts because they are always slimmer than their boundary annotations. To refine the coarse proposals, we propose a boundary transformer module to regress per-vertex offsets from coarse boundary proposal to actual text boundary, as shown in Fig.~\ref{fig:introduction} (c). The boundary transformer module adopts an encoder-decoder structure, in which multi-layer transformer blocks construct the encoder with residual connection, and the decoder is a simple multi-layer perceptron network (MLP). Under the guidance of prior information, the boundary transformer module will gradually refine the coarse boundary proposals via iterative boundary deformation. In addition, we propose a novel boundary energy loss (BEL), consisting of an energy minimization constraint and an energy monotonically decreasing constraint, to further optimize and stabilize the learning of boundary refinement.

The presented work is an extended version to our preliminary conference version TexBPN~\cite{TextBPN}, and it involves the following features: \textbf{\romannumeral1)} We systematically present a unified coarse-to-fine framework via boundary learning for arbitrary shape text detection. \textbf{\romannumeral2)} We propose a powerful yet efficient boundary transformer module for boundary deformation. The boundary transformer module has a stronger long-distance relationship modeling ability, which makes it more robust to changes in the number of control points. \textbf{\romannumeral3)} We propose a novel boundary energy loss (BEL) for optimizing network training. Using the point matching loss and boundary energy loss together will further stabilize the learning of boundary deformation and improve the robustness to noise in training. \textbf{\romannumeral4)} We additionally adopt the lightweight backbone network (ResNet-18) and enhanced backbone network (ResNet-50 with deformable convolution) to verify the state-of-the-art performance and promising efficiency of our method. \textbf{\romannumeral5)} We conduct experiments on additional two new larger datasets (\ie, ICDAR-ArT and ICDAR-MLT2017) to further verify the performance. We also add more exploration experiments and intuitive comparisons. 

In summary, our main contributions in this paper are summarized into four-fold:

\begin{itemize}
	\item We systematically present a unified coarse-to-fine framework via boundary learning for arbitrary shape text detection, which can accurately and efficiently locate text boundaries without post-processing.
	
	\item We propose an innovative boundary transformer module to adaptively and gradually refine coarse text boundaries in an iterative manner.
	
	\item We propose a novel boundary energy loss (BEL) to further optimize and stabilize the learning of iterative boundary refinement.
	
	\item Extensive experiments on publicly available datasets demonstrate the state-of-the-art performance and promising efficiency of our method.
\end{itemize}

The rest of the paper is organized as follows: Section \ref{Related_Work} overviews the related work. Section \ref{Proposed_Method} elaborates on our work. In Section \ref{Experiments}, we demonstrate some experimental results and analysis. Finally, we conclude our work in Section \ref{Conclusion}.

\begin{figure*}[htbp]
	\begin{center}
		\includegraphics[width=1.0\linewidth]{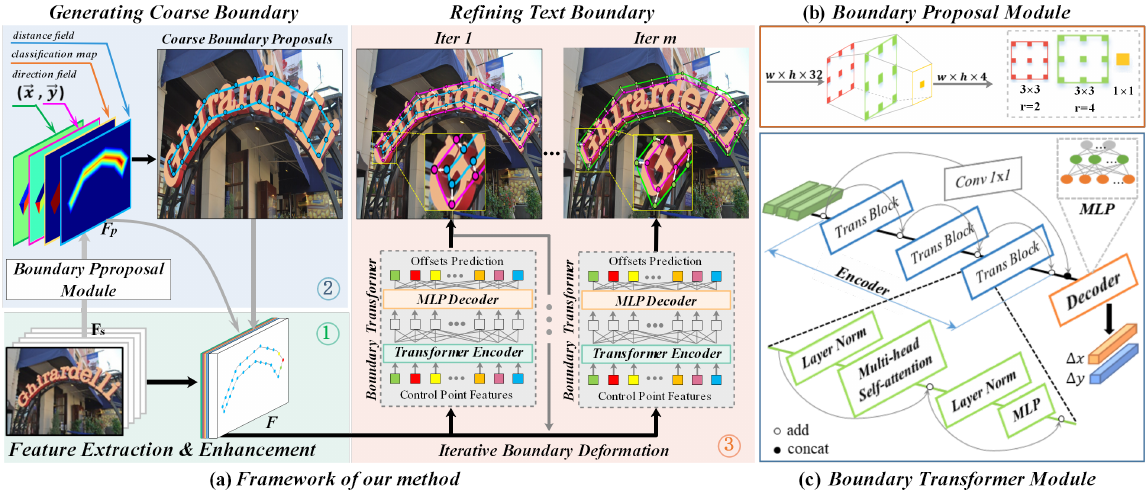}
		\caption{(a)  Framework of our method. Our framework consists of three parts: feature extraction, generating coarse boundary, and refining text boundary. (b) The structure of the boundary proposal module. (c) The structure of the boundary transformer module.}
		\label{fig:framework}
	\end{center}
	\vspace{-1.8em}
\end{figure*} 


\section{Related Work} \label{Related_Work}
\subsection{Regression-based Methods} 
The regression-based methods~\cite{RRPN,textboxes++,RRD,EAST, CVPR19_LOMO,HAM, ZhuLLS21,Zhang2022GraphFN} always modify box-regression based object detection frameworks with word-level and line-level prior knowledge for text instance. However, scene texts often suffer from arbitrary orientation with various aspect ratios. To address this problem, TextBoxes~\cite{textboxes} and TextBoxes++~\cite{textboxes++} use
a series of anchors with different aspect ratios for covering texts with varied lengths. Different from these anchor-based methods, anchor-free methods (\eg, EAST~\cite{EAST}, MOST~\cite{MOST}, and DDR~\cite{DDR}) directly regress the offsets from boundaries or vertexes to the current point for detecting texts. LOMO~\cite{CVPR19_LOMO} introduces an iterative refinement module to iteratively refine the text localization of a direct regression based on bounding box proposals. HAM~\cite{HAM} designs a hidden anchor mechanism to integrate the advantages of the anchor-based method into the anchor-free method. Although regression-based methods have achieved good performance in quadrilateral text detection, they often can't well adapt to arbitrary shape text detection.

\vspace{-0.5em}
\subsection{Connected Component-based Methods}	
The Connected Component (CC) based methods~\cite{Yin-M,SegLink++,TextSnake, CRAFT, DRRG} usually detect individual text parts or characters firstly, followed by a link or group post-processing procedure for generating final texts. SegLink++~\cite{SegLink++} adopts instance-aware component grouping with the minimum spanning tree to achieve arbitrary shape text detection. CRAFT~\cite{CRAFT} detects character-level text and explores the affinity between characters to achieve the final detection. TextDragon \cite{TextDragon} simply groups the detected local text region by their geometric relations. Zhang \etal~\cite{DRRG} adopted CNN to predict the geometry attributes of text components and introduced Graph Convolutional Network
(GCN) to learn and infer the linkage relationships between different text components. Although the flexible representation of CC-based methods is more adapted to arbitrary shape text detection, the detection performance and efficiency usually suffer from the complex post-processing for text components clustering.

\vspace{-0.5em}
\subsection{Segmentation-based Methods}
The segmentation-based methods~\cite{CVPR19_PSENet, TextField, PSENet_v2, DB, MaskTextSpotter, KPN, AAM, ZhuLZLLX19} are mainly inspired by semantic segmentation and achieve text detection by implicitly encoding text instances with pixel masks. In PSENet~\cite{CVPR19_PSENet}, a progressive scale expansion algorithm is applied to 
expand the pre-defined kernels for fusing different-scale segmentation maps. PAN~\cite{PSENet_v2} and LSAE~\cite{CVPR19_LSA}  pull the pixel embedding of the same text and push the pixel embedding of different text by learning the embedding vectors of pixels. TextField~\cite{TextField} adopts a deep direction field to link neighbor pixels and generate candidate text instances. DB~\cite{DB} simplifies the post-processing of text detection in a segmentation network with differentiable binarization. In these methods, segmentation accuracy significantly determines the quality of detected boundaries. 

\begin{figure}[tbp]
	\begin{center}
		\includegraphics[width=0.96\linewidth]{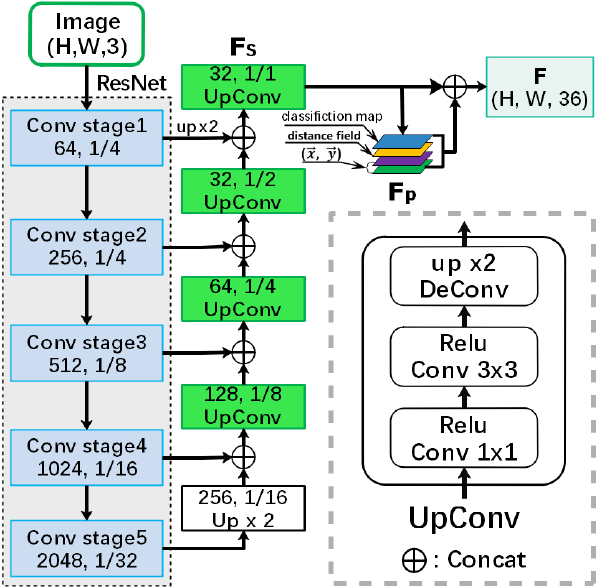}
		\caption{Architecture of our backbone for feature extraction. $ F_{s} $ denotes the shared features and $ F_p $ denotes the prior information (\ie, classification map, distance field, and direction field).}	
		\label{fig:backbone}
	\end{center}%
	\vspace{-1.8em}
\end{figure}

\begin{figure*}[tbp]
	\begin{center}
		\includegraphics[width=0.85\linewidth]{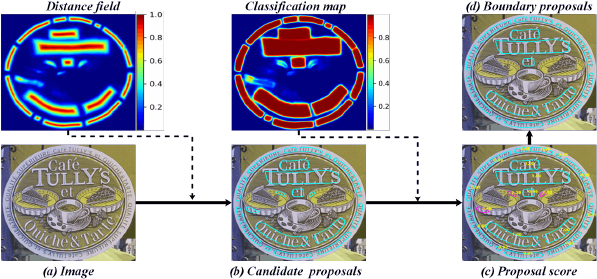}
		\caption{Illustration of boundary proposal generation. We firstly generate candidate boundary proposals by applying a threshold ($ th_d $) to binarize the distance field, and then filter the candidate boundary proposals with a low confidence score.}
		\label{fig:proposal}
	\end{center}%
	\vspace{-1.6em}
\end{figure*}

\vspace{-0.5em}
\subsection{Contour-based methods} 
Contour-based~\cite{CVPR19_ATRR,ContourNet, ABCNet, FCENet, TextRay, PCR} try to directly model the text boundary for detecting the arbitrary shape text. ABCNet~\cite{ABCNet} and FCENet~\cite{FCENet} model text instance contours with curve modeling (Bezier-Curve and Fourier-Curve), which can fit any closed contour with progressive approximation. TextRay~\cite{TextRay} formulates the text contours in the polar
system and proposes a single-shot anchor-free framework to predict geometric parameters and output simple polygon detections. PCR~\cite{PCR} proposes a progressive contour regression approach based top-down detection framework to detect arbitrary-shape scene texts. Similar to PCR~\cite{PCR}, some other methods~\cite{CVPR19_ATRR,ContourNet} also adopt a top-down detection framework, which regresses the key points on text contours within the text proposals. Compared with segmentation-based methods, 
there is still much space for exploration in performance and speed.


\section{Proposed Method} \label{Proposed_Method}
\subsection{Overview}
As illustrated in Fig.~\ref{fig:framework}, our method presents a unified coarse-to-fine framework, which mainly consists of a feature extraction backbone, a boundary proposal module, and an iteratively optimized boundary transformer module. The architecture of our backbone for feature extraction is illustrated in Fig.~\ref{fig:backbone}. To preserve spatial resolution and take full advantage of multi-level information, we exploit a multi-level
feature fusion strategy (similar to FPN~\cite{FPN}). The boundary proposal module composed of multi-layer dilated convolutions uses the shared features for performing text pixels classification, generating the distance field and direction field~\cite{TextField}. Then, we use this information to produce coarse boundary proposals. Each boundary proposal consists of $ N $ control points, representing a possible text instance. For refining the coarse proposals, we perform iterative boundary deformation with a boundary transformer module to refine text boundaries under the guidance of prior information.

\subsection{Boundary Proposal}
The boundary proposal module consists of multi-layer dilated convolutions, including two $ 3\times 3 $ convolution layers with different dilation rates and one $ 1\times 1 $ convolution layer, as shown in Fig.~\ref{fig:framework} (b). It will use the shared features extracted from the backbone network to generate a classification map, distance field map, and direction field map. 

Similar to other text detection methods~\cite{EAST, TextSnake,DRRG}, the classification map contains the classification confidence of each pixel (text/non-text). As in~\cite{TextField, Super_BPD}, the direction field map ($ V $) consists of a two-dimensional unit vector ($ \vec{x}
,\vec{y}$), which indicates the direction of each text pixel inside the boundary to its nearest pixel on the boundary (pixel-to-boundary). For each pixel $ p_{i} $ inside a text instance $ T $ ,  we will find its nearest pixel $ b_{i} $ on the text boundary. Then, a two-dimensional unit vector $ \mathcal{V}(P_{i}) $ that points away from
the text pixel $ p_i $ to $ b_i $ can be formulated as 
\begin{equation}
	\mathcal{V}(p_{i}) \; = \; \left\{\begin{matrix}\ \overrightarrow{p_{i}  b_{i} }/\left\vert\overrightarrow{p_{i}  b_{i}}\right\vert, & p_i\in T
		\\ \\
		(0,0), & p_i \not\in T \end{matrix}\right.\label{eq:dir}
\end{equation}
where $\left\vert\overrightarrow{b_i p_i}\right\vert$ represents the distance between $ b_i $ and text pixel $ p_i $. For the non-text area ($ p_i \not\in T $), we represent those pixels with $(0,0)$.  The unit vector $ \mathcal{V}(p_i) $ directly encodes the approximate relative location of $ p_i $ inside $ T $ meanwhile highlighting the boundary between adjacent text instances~\cite{TextField}. In addition, it provides direction information for boundary deformation.

For boundary deformation, the relative position distance information is as important as the direction information. In this work, the distance field map ($ \mathcal{D} $) 
is a normalized distance map, and the normalized distance of the text pixel $ p_i $ to
nearest pixel $ b_i $ on the text boundary is defined as
\begin{equation}
	\mathcal{D}(p_i) \; = \; \left\{\begin{matrix}\ \left\vert\overrightarrow{p_i b_i}\right\vert/L, & p_i\in T
		\\ \\
		0, & p_i \not\in T \end{matrix}\right.
\end{equation}
For the non-text area ($ p_i \not\in T $), we represent the distance of those pixels with $0$. $ L $ represents the scale of text instance $ T $ where the pixel $ p_i $ is located, and is defined as
\begin{equation}
	L = max(\{ \scriptstyle \left\vert\overrightarrow{p_i b_i}\right\vert \displaystyle \big| p_i\in T\})
\end{equation}
where $ L $ is maximum distance of the text pixel $ p_i $ to the nearest boundary pixel $ b_i $  
in text instance $ T $. $ \mathcal{D}(p_i) $ can directly encodes the relative distance of $ p_i $ inside $ T $ and further highlights the boundary between adjacent text instances. Besides, it provides relative distance information for boundary deformation.

With the distance field map ($ \mathcal{D} $), we firstly generate candidate boundary
proposals by applying a threshold ($ th_d $) to binarize the distance field, as shown in Fig.~\ref{fig:proposal} (b). However, these candidate boundary proposals inevitably contain false detections. Hence, we calculate the confidence score ($ th_s $) for every candidate boundary proposal according to the classification map. Finally, we gain the coarse boundary proposals by removing candidate proposals with low confidence scores ($ th_s $), as shown in Fig.~\ref{fig:proposal} (c) and (d).

\subsection{Boundary Transformer}
In our method, we perform arbitrary shape text detection by
transforming the coarse boundary proposals to accurate text boundaries. Specifically, we learn to predict per-vertex offsets pointing to the text boundary based on the coarse boundary proposals in an iterative manner. For each coarse boundary proposal represented as a closed polygon, we will uniformly sample $ N  $ control points for facilitating batch processing, as shown in~Fig.~\ref{fig:introduction}. The sequence of these control points not only contains sequence context but also contains topological context (\eg, shape and spatial distribution). To fully utilize and excavate these two contexts for refining coarse text boundaries, we propose a boundary transformer module to efficiently perform feature learning and predict accurate per-vertex offsets pointing to the text boundary.

Let $ c_i = [x_i, y_i]^{T} $ denote the location of the $ i$-th control point, and $ \textbf{P} = \{{c}_{_0},...,{c}_{_i},...,{c}_{_{N-1}}\} $ be the set of all control points. For a boundary proposal with $ N  $ control points, we first construct feature vectors for each control point. The input feature $ f_{i} $ for a control point $ c_i $ is a concatenation of  32-D shared features $ F_s $ and 4-D prior features $ F_p $ (\eg, classification map, distance field, and direction field). Therefore, the features of a control point are extracted from the corresponding location in $ F:f_{i} = concat\{F_s(x_i,y_i), F_p(x_i,y_i)\}$. Here, $ F_s(x_i,y_i) $ and  $ F_p(x_i,y_i) $ are computed by bilinear interpolation.

After obtaining the feature matrix of the boundary proposal,  a novel boundary transformer module is adopted to efficiently perform feature learning and refine the coarse boundary proposals iteratively. The boundary transformer module adopts an encoder-decoder structure, in which the encoder consists of three layers with transformer blocks equipped with residual connections, and the decoder is a simple multi-layer perceptron network (MLP), as shown in Fig.~\ref{fig:framework} (c). The encoder will encode the feature map ($ B \times N \times 36$) of boundary proposals into an embedded feature map ($ B \times N \times 128$). Each encoder layer can be formulated as
\begin{equation}
	X^{'} = X \oplus TransBlock(X)
\end{equation}
where $ X $ (size: $ N \times C $) denotes the feature matrix of boundary proposal; ``$ \oplus $'' refers to the add operation. Each transformer block has a standard architecture and consists of a multi-head self-attention and a multi-layer perceptron network (MLP), as shown in Fig.~\ref{fig:framework} (c). In our method, we need to learn relative per-vertex offset, so the absolute position encodings based on the pixel position in an image or the sequence order of the control point in the boundary proposal are unnecessary. Our extensive experiments also indicate that position encodings can not improve the accuracy of detected boundaries. Therefore, we remove position encodings in our boundary transformer.

\begin{figure}[tbp]
	\begin{center}
		\includegraphics[width=0.98\linewidth]{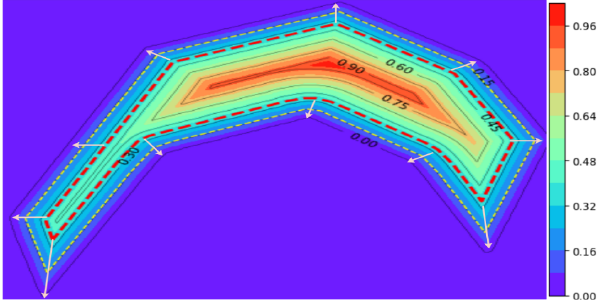}
		\caption{Illustration of the energy field. In our method, we regard the distance	field as our energy field. Therefore, the energy of text pixels is always greater than 0, and the energy of background pixels is 0. The maximum energy value is 1 for each text instance. Boundary deformation aims to search the circle of pixels with the lowest energy in the text. The red dotted line denotes a coarse boundary proposal.}
		\label{fig:energy}
	\end{center}%
	\vspace{-1.2em}
\end{figure}

The decoder of our boundary transformer module consists of a three-layer perceptron network together with   $ 1 \times 1 $ convolutions activated by ReLU. It will learn to predict offsets ($ Y $) between control points and target points. Mathematically, the prediction of offsets $ Y $ can be expressed as
\begin{equation}
	Y = Decoder(Conv_{1 \times 1}(X) || X^{'''})
\end{equation}
where $ Conv_{1 \times 1} $ denotes one-layer $ 1 \times 1 $ convolution layers with 128 dimensions; operator $ || $ represents matrix concatenation along the feature dimension; $ X^{'''} $ represents the feature map after three encoder layers;  $ Y $ is the sequence of the per-vertex offsets for boundary proposals. To refine the text boundaries, we perform boundary deformation iteratively with the boundary transformer module, as shown in Fig.~\ref{fig:framework}. Notably, the per-vertex predicted offsets are limited to no more than 16-pixel distance in each iteration for ensuring the convergence of model training.

\subsection{Boundary Energy Loss}
Inspired by the active contour model~\cite{Active_Contour}, we propose a novel boundary energy loss ($ \mathcal{L}_{E}  $) to minimize boundary energy for further optimizing and stabilizing the learning of boundary deformation. The boundary energy loss includes an energy minimization constraint ($ \mathcal{L}_{be}  $) for individual boundary deformation and an energy monotonically decreasing constraint ($ \mathcal{L}_{ie}  $) for boundary iterative refinement. Hence, the boundary energy loss ($ \mathcal{L}_{E}  $) is the sum of  $ \mathcal{L}_{be}  $ and $ \mathcal{L}_{ie}  $, as
\begin{equation}
	\mathcal{L}_{E} = \mathcal{L}_{be} + \mathcal{L}_{ie},
\end{equation}

To avoid additional parameters, in our method, we regard the distance field as our energy field, as shown in Fig.~\ref{fig:energy}. Therefore, the energy of the control point $ c_i  $ is equal to its corresponding distance value in the distance field, \ie $ E(c_i) = D(c_i) $. The energy of a boundary is defined as the sum of the energy of all control points in the boundary, as 
\begin{equation}
	E(\textbf{P}) = \sum_{i=1}^{N}E(c_i),
\end{equation}

The energy minimization constraint ($ \mathcal{L}_{be}  $) enforces the boundary optimization along the energy-decreasing direction in training. $ \mathcal{L}_{be}  $ is defined as
\begin{equation}
	\mathcal{L}_{be} =\frac{1}{\mathbb{T}}\sum_{\textbf{P} \in \mathbb{T}}E(\textbf{P})=\frac{1}{\mathbb{T}}\sum_{P \in \mathbb{T}} \sum_{i=1}^{N}E(c_i),
\end{equation}

The energy monotonically decreasing constraint ($ \mathcal{L}_{ie} $) the energy value of the predicted boundary related to the current iteration to be lower than the previous one, thus forcing the prediction to higher accuracy. The $ \mathcal{L}_{ie}  $ is  defined as
\begin{equation}
	\mathcal{L}_{ie} =\frac{1}{\mathbb{T}}\sum_{\textbf{P} \in \mathbb{T}}max(0,E(\textbf{P})^{i} - E(\textbf{P})^{i-1}),
\end{equation}
where $ E(\textbf{P})^{i} $ denotes the energy value of the predicted boundary related to the current iteration and $ E(\textbf{P})^{i-1} $ denotes the energy value of predicted boundary related to the previous iteration. $ E(\textbf{P})^{0} $ is the energy value of boundary proposal.

\subsection{Optimization}
\noindent The total loss $ \mathcal{L} $ of our method can be formulated as
\begin{equation}
	\mathcal{L} = \mathcal{L}_{BP} + \dfrac{\lambda * \mathcal{L}_{BT}}{1+e^{(i-eps)/eps}},\label{tal_loss}
\end{equation}
where $ \mathcal{L}_{BP} $ is the loss for the boundary proposal,  and $ \mathcal{L}_{BT} $ is the loss for the adaptive boundary transformer; $ eps $ denotes the maximum epoch of training, and $ i $ denote the $ i$-th epoch in training. In our experiments, $ \lambda  $ is set to 0.1. In Eq.~\ref{tal_loss}, $ \mathcal{L}_{BP} $ is computed as
\begin{equation}
	\mathcal{L}_{BP} = \mathcal{L}_{cls} + \alpha* \mathcal{L_{D}}+ \beta*\mathcal{L_{V}},
\end{equation}
where $ \mathcal{L}_{cls} $ is a cross-entropy loss for pixels classification, and $ \mathcal{L_{D}} $ is a $ L_{2} $ loss for distance field. OHEM~\cite{OHEM} is adopted for $ \mathcal{L_{D}} $ in which the ratio between the negatives and positives is set to 3:1.  To balance the losses in $ \mathcal{L}_{BP} $, the weights $ \alpha $ is set to 3 and the weights $ \beta $ is set to 0.5. 
Similar to~\cite{Super_BPD}, $ \mathcal{L_{V}} $  consists of $ L_2 $-norm distance and angle distance for direction field $ \mathcal{V} $:
\begin{equation}
	\mathcal{L_{V}} = \sum_{p \in \Omega} {  w(p){\|\mathcal{V}_p - \hat{\mathcal{V}}_p\|}_2 } + \frac{1}{\mathbb{T}}\sum_{p \in \mathbb{T}} (1 - \cos( \mathcal{V}_p, \hat{\mathcal{V}}_p)) ,
	\label{eq:finalloss}
\end{equation}
where $ \Omega $ represents the image domain; the weight ($w(p) = 1/\sqrt{|GT_p|}$) at pixel $p$ is proportional to the inverse square root of the size of ground truth segment $GT_p$ containing $p$.

$\mathcal{L}_{BT} $ consists of a point matching loss $ \mathcal{L}_{P} $ and a boundary energy loss $ \mathcal{L}_{E} $ .  In our method, the text boundary will be iteratively refined. Hence, the total loss of boundary transformer ( $\mathcal{L}_{BT} $) is computed as
\begin{equation}
	\mathcal{L}_{BT} = \frac{1}{m}\sum_{i=1}^{m}(\mathcal{L}_{E}(i) + \mathcal{L}_{P}(i)),
\end{equation}
where $ m $ denotes the number of iteration;  $ \mathcal{L}_{P}(i) $ and $ \mathcal{L}_{E}(i) $  donate the point matching loss and boundary energy loss  in i-$ th $ iteration, respectively.

In our method, the predictions ($ \textbf{P} = \{{c}_{_0},...,{c}_{_i},...,{c}_{_{N-1}}\} $) and ground-truths ($\hat{\textbf{P}} = \{{\hat{c}}_{_0},...,{\hat{c}}_{_i},...,{\hat{c}}_{_{N-1}}\} $) of control point have equal size and same order (counter-clockwise).  Similar to~\cite{Curve_GCN}, the matching
loss for $ \textbf{P} $ and $ \hat{\textbf{P}}$ is defined as
\begin{equation}
	\mathcal{L}(\textbf{P}, \hat{\textbf{P}}) = \min_{j \in [1\cdots, N]}\sum_{i=1}^{N}{smooth_{L1}(c_{_i}, {\hat{c}}_{_{(j+i)\%N}})},
\end{equation}
where $N$ is the number of control points. Hence, the point matching loss ($\mathcal{L}_{P} $) is defined as
\begin{equation}
	\mathcal{L}_{P} = \frac{1}{\mathbb{T}}\sum_{p \in \mathbb{T}} \mathcal{L}(\textbf{P}, \hat{\textbf{P}}),
\end{equation}
where $\mathbb{T}$ represents all text instances in an image; $  \textbf{P} $ represents the control point set for a text instance $ T $ ($ T \in  \mathbb{T} $).


\section{Experiments} \label{Experiments}
\subsection{Datasets}
\noindent\textbf{Total-Text}: It consists of $1,255$ training and $300$ complex testing images, including horizontal, multi-oriented, and curved text with polygon and word-level annotations.

\noindent\textbf{CTW-1500}: It consists of $1,000$ training and $500$ testing images, and curved text instances are annotated by polygons with 14 vertices.

\noindent\textbf{MSRA-TD500}: It consists of $500$ training and $200$ testing images, including English and Chinese texts, containing multi-lingual long texts with multi-orientations.

\noindent\textbf{ICDAR-ArT}: It is a large-scale multi-lingual arbitrary-shape scene text detection dataset, including 5,603 training images and 4,563 testing images. The text regions are annotated by the polygons with an adaptive number of key points.

\noindent\textbf{ICDAR-MLT17}: It consists of $7,200$ training images, $1,800$ validation images, and $9,000$ test images with multi-lingual ($9$ languages) texts annotated by quadrangle.

\noindent\textbf{SynthText}: It contains 800k synthetic images generated by blending natural images with artificial text, which are all word-level annotated.

\begin{table*}[htbp]
	\begin{center}
		\renewcommand{\arraystretch}{1.3}
		\caption{Ablation experiments for deformation model on Total-Text and CTW-1500. The best score is highlighted in \textbf{bold}.}	\label{table:Ablation}
		\begin{tabular}{||c||c|c|c|c||c|c|c|c||}
			\hline
			\multicolumn{1}{||c||}{ \multirow{2}*{ \textbf{Methods}}}
			& \multicolumn{4}{c||}{\textbf{   Total-Text}} 
			& \multicolumn{4}{c||}{\textbf{ CTW-1500}}\\
			\cline{2-9}
			&\textbf{Recall}
			& \textbf{Precision} & \textbf{F-measure}&
			\textbf{FPS}&
			\textbf{Recall}& 
			\textbf{Precision} & \textbf{F-measure}&
			\textbf{FPS}\\
			\hline
			Fully connected networks (FC) &81.56 &90.16 &85.65&9.5&78.32 &85.03 &81.54&11.1\\
			Recurrent neural network (RNN) &83.31 &87.71 &85.93&11.2&81.26 &86.00 &83.56&12.2\\
			Circular convolution network (CCN) &82.80 &89.73 &86.13 &9.3&80.35 &84.88 &82.55&10.9 \\
			Graph convolution network (GCN)&82.74 &89.94 &86.19  &10.4 &80.31&86.12 &83.12&11.9\\
			\textbf{Adaptive deformation (AD)}&83.30 &90.76 &86.87 &10.6&80.57 &87.66 &83.97&12.1 \\
			\textbf{Boundary transformer (BT)}  &83.88 &\textbf{90.81} &87.21  &\textbf{12.3}&81.07&87.58 &84.20 &\textbf{14.7}\\ 
			\textbf{Boundary transformer (BT)}+BEL &\textbf{85.29} &89.86 &\textbf{87.52} &12.0&\textbf{81.12}&\textbf{88.08} &\textbf{84.46} &\textbf{14.7}\\ 
			\hline
		\end{tabular}
	\end{center}%
	\vspace{-1.5em}
\end{table*}

\subsection{Implementation Details}\label{trainstep}
The ResNet~\cite{ResNet} is adopted as a backbone. In our experiments, we randomly crop the text region and resize them to $ 640 \times 640 $ for training the model with 660 epochs (except MSRA-TD500 with 1200 epochs). The mini-batch is set to 12 for ``ResNet-50-1s'', and 48 for ``ResNet-18-4s''.  The Adam~\cite{ADAM} is applied as our optimizer. Without pre-training, the initial learning rate is set to $ 0.001 $ and decayed by $ 0.9 $ after every 50 epochs. In some experiments, we fine-tune our models on the ICDAR-MLT17 model for more fair and comprehensive comparisons. In fine-tuning, the initial learning rate is $ 0.0001 $ and will be decayed by $ 0.9 $ after every 50 epochs. The data augmentation includes random rotation with an angle (sampled by Gaussian distribution in ($-30^{\circ}, 30^{\circ}$)), random cropping, and random flipping. In inference, we keep the aspect ratio of test images, then resize and pad them into the same size for testing.  The code is implemented with PyTorch 1.7 and Python 3. Training is performed on a single GPU (RTX-3090), and testing is performed on a single GPU (GeForce RTX-2080) with Intel Xeon Silver 4108 CPU @ 1.80GHz.

Notably, ``Ext'' denotes extra data is used for pre-training, ``Syn'' denotes SynthText, ``MLT'' denotes ICDAR-MLT17, ``MLT$ ^{+} $'' denotes ICDAR-MLT17 together with extra datasets, ``ArT$ ^{-} $'' denotes it selects dataset from ICDAR-ArT (excluding the test set of Total-Text or CTW-1500). ``R", ``P", and ``F" represent Recall, Precision, and F-measure, respectively. For a fair comparison of detection speed, similar to DB~\cite{DB}, the speed is evaluated by performing a testing image 50 times to exclude extra IO time.

\subsection{Exploration Experiment}\label{exp_ablation_aam}
In exploration experiments, we only train the model on corresponding real-world datasets by 660 epochs without any pre-training. Adam~\cite{ADAM} is adopted as the optimizer with an initial learning rate of $ 0.001 $. The other training settings are identical to the fine-tuning process in Sec~\ref{trainstep}. During testing, we set both sides of the image in the range of (640, 1024) while keeping its aspect ratio. Then, we pad the test images to 1024$  \times $1024 and send them to the network for processing. In all exploration experiments, We use a fixed threshold on the same dataset, \ie, Total-Text ($ th_d $=0.3, $ th_s $=0.85) and CTW-1500 ($ th_d $=0.35, $ th_s $=0.825). The number of iterations is set to 3 by default in our experiments.

\noindent\textbf{Effectiveness of boundary transformer module.}
To verify the effectiveness of the proposed boundary transformer module, we conducted some exploration experiments on Total-Text and CTW-1500. Unless BEL is specially marked, the boundary energy loss (BEL) is not used during training. For a fair comparison, we use the same encoder with a three layers MLP activated by Relu. Then, we conduct a comparative experiment with boundary deformation module, \ie, fully connected networks (FC), recurrent neural network (RNN), circular convolution network (CCN)~\cite{DeepSnake}, graph convolution network (GCN), and adaptive deformation (AD) module, 
to compare its performance with the proposed boundary transformer (BT) module. As listed in Tab.$ \, $\ref{table:Ablation}, our boundary transformer module achieves the best performance compared with the other five methods both on Total-Text and CTW-1500. Specifically, the boundary transformer module respectively achieves performance improvement by 1.02\% in terms of F-measure on Total-Text and by 1.08\% in terms of F-measure on CTW-1500 against GCN. Compared with the adaptive deformation module~\cite{TextBPN}, the proposed boundary transformer (BT) also achieved consistent performance improvement on Total-Text and CTW-1500. Especially our boundary transformer module respectively outperforms the adaptive deformation module by 0.58\% on Total-Text, and 0.5\% on CTW-1500 in terms of recall. In addition, our boundary transformer can achieve promising efficiency due to its lightweight design.

\begin{figure}[tbp]
	\begin{center}
		\includegraphics[width=0.98\linewidth]{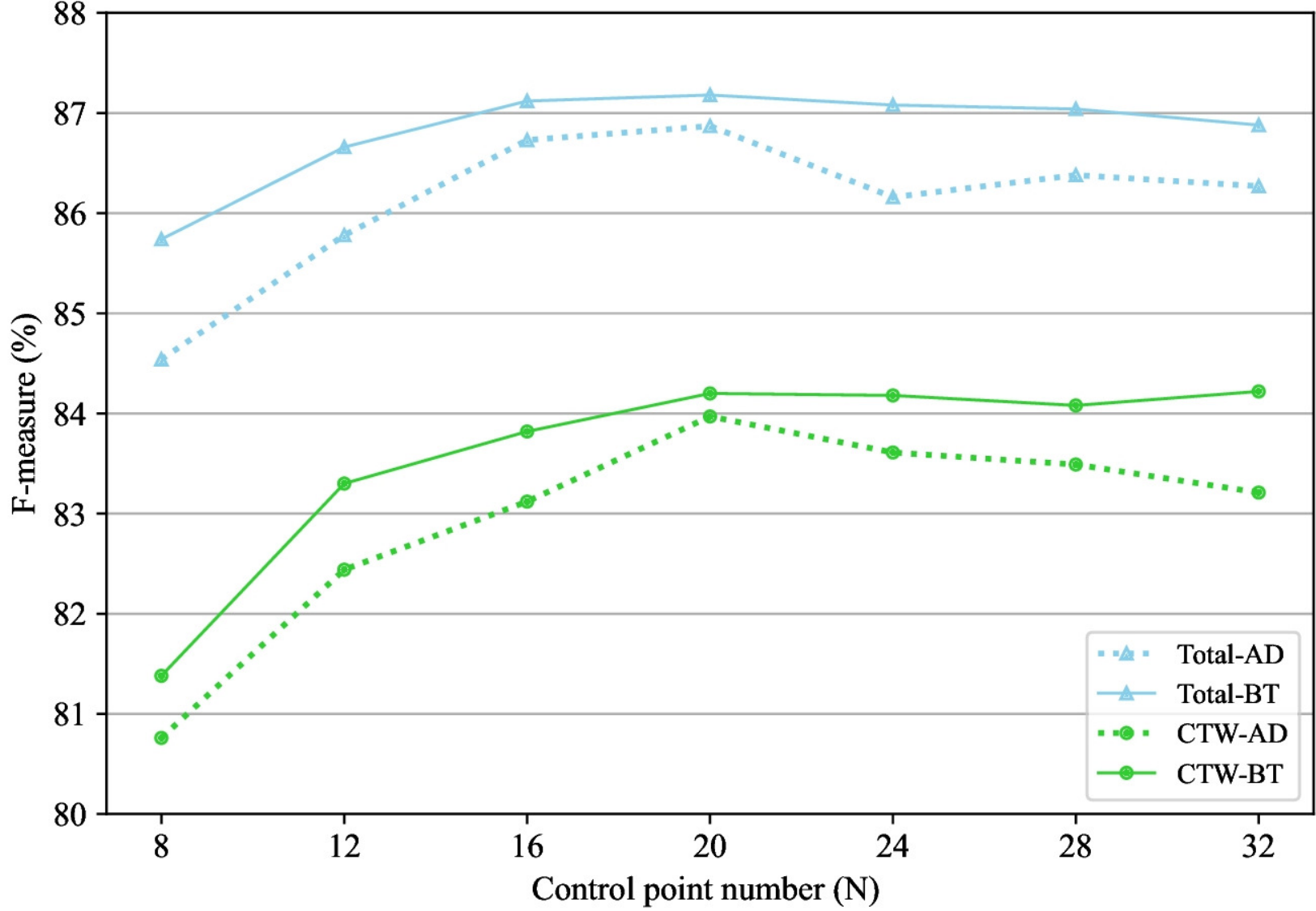}
		\caption{Experimental results of the control point number (N). The blue solid line represents the experimental results of the boundary transformer (BT) on Total-Text, and the blue dotted line represents the experimental results of the adaptive deformation (AD) on Total-Text. The green solid line represents the experimental results of the boundary transformer (BT) on CWT-1500, and the green dotted line represents the experimental results of the adaptive deformation (AD) on CWT-1500.}
		\label{fig:ctr_pts}
	\end{center}
	\vspace{-1.8em}
\end{figure}

\begin{figure*}[htbp]
	\subfigcapskip=-3pt
	\centering
	\includegraphics[width=1.0\linewidth]{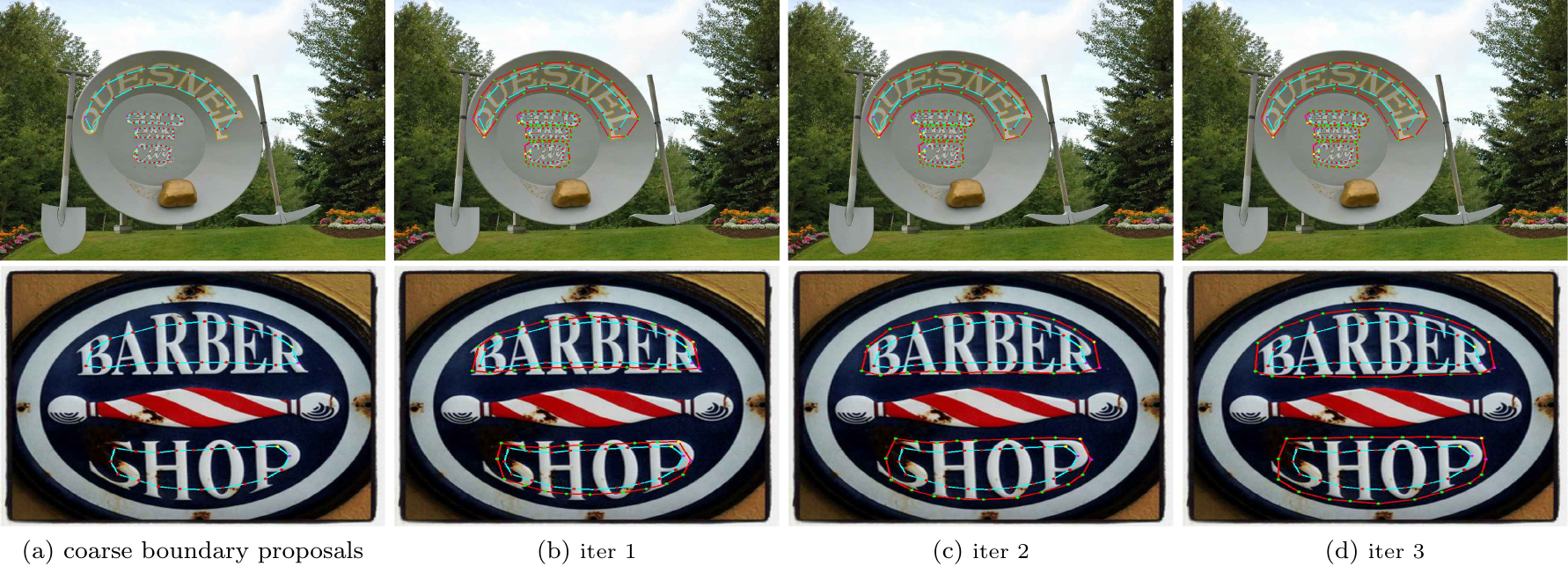}
	\centering
	\caption{ Visual results of different iterations. The blue contours are coarse boundary proposals, and the red contours are detection boundaries after iterative boundary transformer.}
	\label{fig:iter_fig}
	\vspace{-1.6em}
\end{figure*}

\noindent\textbf{Influence of control point number ($ N $).}
Here, we conduct experiments without the help of boundary energy loss to verify the influence of the different control point numbers ($ N $). The control point number will range from 8 to 32 with an interval of 4. As shown in Fig.~\ref{fig:ctr_pts}, we can find that the score of F-measure grows rapidly with the increase of the control point number in the low range. Especially, too few control points will cause great degradation of performance. This is because the detection boundary often can’t correctly cover the whole text when the control number is too small. When $ N $ reaches a relatively high value (\eg, 20), the detection performance of the proposed boundary transformer tends to be stable on both Total-Text (87\%$ ^{+} $ F-measure) and CTW-1500 (84\%$ ^{+} $ F-measure). 
In contrast,  the detection performance of the adaptive deformation module~\cite{TextBPN} tends to drop. The increase of the control point number means that the sampling spacing is reduced and the adjacent control points are closer in the image. However, the adaptive deformation module that combines the GCN and RNN adopts a fixed adjacency matrix, which will make its receptive field smaller while the control points become dense. Hence, the proposed boundary transformer based on the transformer network has more advantages than the adaptive deformation module for long sequence learning and modeling during boundary deformation. In our other experiments, we set the number of control points to 20 to balance the detection performance and computation cost.

\begin{table}[tbp]
	\begin{center}
		\renewcommand{\arraystretch}{1.1}
		\caption{Ablation study for classification map (\textbf{cls}), distance field (\textbf{dis}), and direction field (\textbf{dir}) on Total-Text.} 
		\label{table:Ablation_field}
		\begin{tabular}{c|ccc|ccc}
			\hline
			Methods&{\textbf{Cls}}&{\textbf{Dis}} & {\textbf{Dir}} & \textbf{Recall}& \textbf{Precision} & \textbf{F-measure} \\
			\hline 
			\multicolumn{1}{c|}{ \multirow{3}*{AD}}&{$ \checkmark $}&{$ \times $} &{$ \times $}  &76.96 &83.01  &79.87\\
			&{$ \checkmark $} &{$ \checkmark $}&{$ \times $} &81.97 &88.95 &85.32\\
			&{$ \checkmark $}&{$ \checkmark $}  &{$ \checkmark $} &\textbf{83.30} &\textbf{90.76} &\textbf{86.87} \\
			\hline
			\hline
			\multicolumn{1}{c|}{ \multirow{3}*{BT}}&{$ \checkmark $}&{$ \times $} &{$ \times $}  &77.11 &82.55  &79.74\\
			&{$ \checkmark $} &{$ \checkmark $}&{$ \times $} &82.37 &89.59 &85.83\\
			&{$ \checkmark $}&{$ \checkmark $}  &{$\checkmark $} &\textbf{83.88} &\textbf{90.81} &\textbf{87.21} \\
			\hline
		\end{tabular}
	\end{center}%
	\vspace{-1.8em}
\end{table}

\noindent\textbf{Influence of prior information.}
We conduct ablation studies on Total-Text to verify the
influence of prior information (\ie, classification map (\textbf{Cls}), distance field (\textbf{Dis}), and direction field (\textbf{Dir})). For a fair comparison, boundary energy loss (BEL) is also not used in training.  The detailed results are listed in Tab.~\ref{table:Ablation_field}. According to Tab.~\ref{table:Ablation_field}, the performance of both our method (BT)and the adaptive deformation module (AD) are unsatisfactory when prior information only includes the classification map.
This is because the classification map can't avoid the adhesion of adjacent text, and the information contained is is insufficient for boundary deformation.
When the distance field is introduced,  the performance will be improved significantly. Specifically, the F-measure is improved by 5.45\% for AD and 6.09\% for BT. The direction field also can bring 1.55\% for AD and 1.38\% for BT in terms of F-measure. In addition, the improvement of recall brought by the direction field is more obvious (1.51\% for BT and 1.33\% for AD). It indicates that the direction field positively affects the expansion of boundary proposals.

\begin{table}[tbp]
	\begin{center}
		\caption{Experimental results of different iterations on CTW-1500.}	\label{table:iteration}
		\begin{tabular}{|m{1.2cm}<{\centering}
				|m{0.445cm}<{\centering}
				m{0.45cm}<{\centering}
				|m{0.45cm}<{\centering}
				m{0.45cm}<{\centering}
				|m{0.45cm}<{\centering}
				m{0.45cm}<{\centering}
				|m{0.45cm}<{\centering}
				m{0.45cm}<{\centering}|}
			\hline
			\multicolumn{1}{|c|}{ \multirow{2}*{Methods}}
			& \multicolumn{2}{c|}{\textbf{  Iter. 1 }} 
			& \multicolumn{2}{c|}{\textbf{ Iter. 2}}
			&\multicolumn{2}{c|}{\textbf{ Iter. 3}}
			&\multicolumn{2}{c|}{\textbf{ Iter. 4}}\\
			\cline{2-9} 
			&F&FPS
			&F&FPS
			&F&FPS
			&F&FPS\\
			\hline
			\textbf{AD}& 82.24 &13.7  & 83.33 & 12.8 & 83.97 & 12.1 &84.06 & 11.6\\
			\textbf{BT} &82.72 & 16.0  &83.63 &15.3 &84.20 & 14.7 &84.30 & 14.0 \\ 
			\textbf{BT}(+BEL)&83.34 & 15.8 &83.95 &15.2 &84.46 & 14.7 &84.60 & 14.0\\ 
			\hline
		\end{tabular}
		\vspace{-1.8em}
	\end{center}%
\end{table}

\noindent\textbf{Influence of iteration number.}
To fully validate the influence of iteration number, we visualize the intermediate results and further compare the detection performance with different inference iterations. As shown in Fig.~\ref{fig:iter_fig}, the detection boundaries become more accurate along with the increase of iterations. As listed in Tab.~\ref{table:iteration}, with the increase in the number of iterations, the detection performance is gradually improved, but the inference speed drops gradually. When the number of iterations increases from 3 to 4, the performance improvement is not very obvious. Considering the balance of efficiency and performance, the number of iterations is set to 3 by default in our experiments. Compared with AD, BT has better detection performance in all iterations. In addition, BEL can bring significant improvement in detection performance with fewer iterations. By referring to Tab.~\ref{table:iteration}, we observe that each additional iteration using BT incurs a cost of merely 2.5 ms, while the backbone consumes around 60 ms. Hence, even after iterating three times, the total time required is only 7.5 ms, which is significantly lower than the time needed for the complicated post-processing of other methods like PSENet or TextPMs. Specifically, the post-processing in PSENet~\cite{CVPR19_PSENet} and TextPMs~\cite{TextPMs} takes roughly 75 ms. Therefore, our iterative refinement is remarkably more efficient compared to these post-processing methods.

\begin{table}[tbp]
	\begin{center}
		\renewcommand{\arraystretch}{1.1}
		\caption{Ablation study for boundary energy loss (BLE) on Total-Text.} 
		\label{table:Ablation_BEL}
		\begin{tabular}{c|cc|ccc}
			\hline
			Methods&{\textbf{Dir}} & {\textbf{BEL}} & \textbf{Recall}& \textbf{Precision} & \textbf{F-measure} \\
			\hline 
			\multicolumn{1}{c|}{ \multirow{4}*{BT}}
			&{$ \times $}&{$ \times $} &82.37 &89.59 &85.83\\
			&{$ \checkmark $}&{$ \times $} &83.88 &\textbf{90.81} &87.21\\
			&{$ \times  $} &{$ \checkmark$} &84.59 &89.82 &87.13\\
			&{$ \checkmark $} &{$\checkmark $} &\textbf{85.29} &89.86 &\textbf{87.52} \\
			\hline
		\end{tabular}
	\end{center}%
	\vspace{-1.5em}
\end{table}

\begin{figure}[tbp]
	\subfigcapskip=-3pt
	\centering
	\begin{center}
		\includegraphics[width=0.99\linewidth]{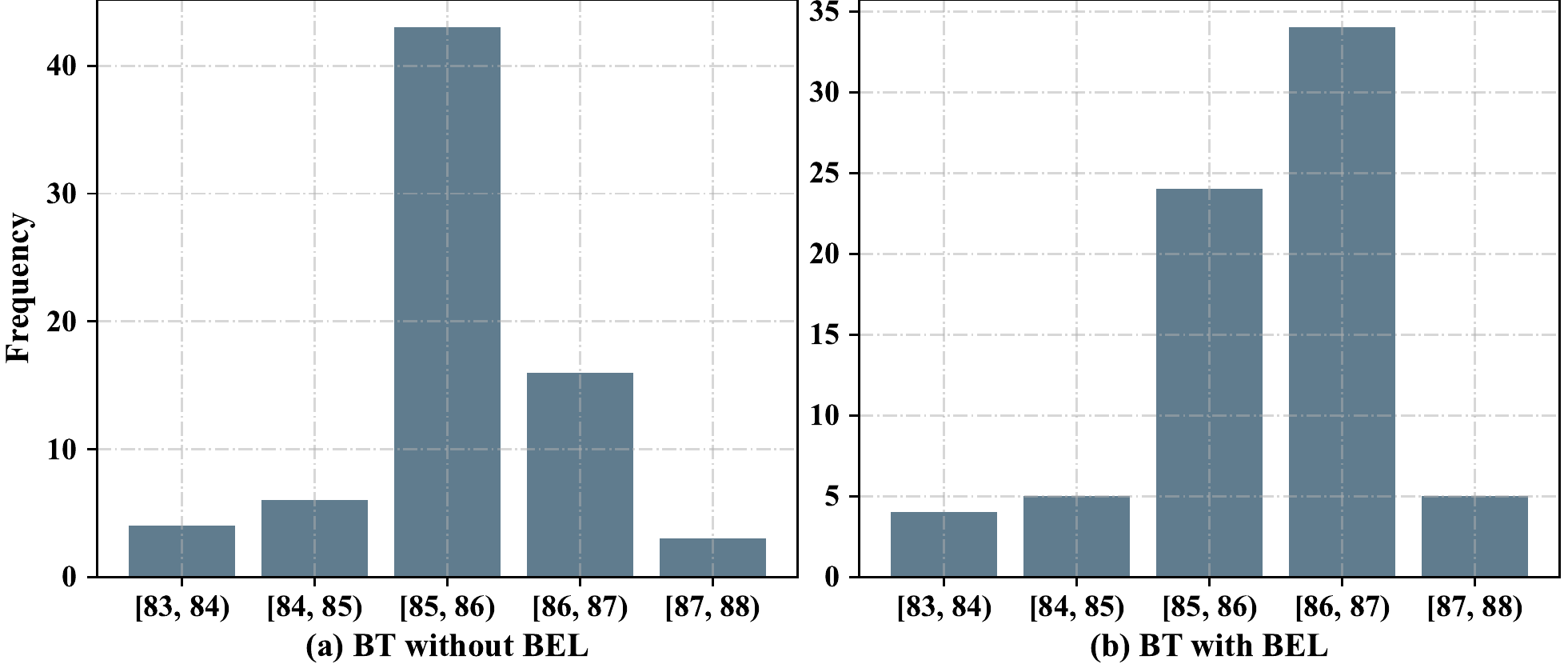}
		\caption{The frequency distribution histogram of detection performance (F-measure) with or without BEL in training.}
		\label{fig:Frequency}
	\end{center}
	\vspace{-1.8em}
\end{figure}

\begin{figure*}[htbp]
	\subfigcapskip=-3pt
	\centering
	\includegraphics[width=1.0\linewidth]{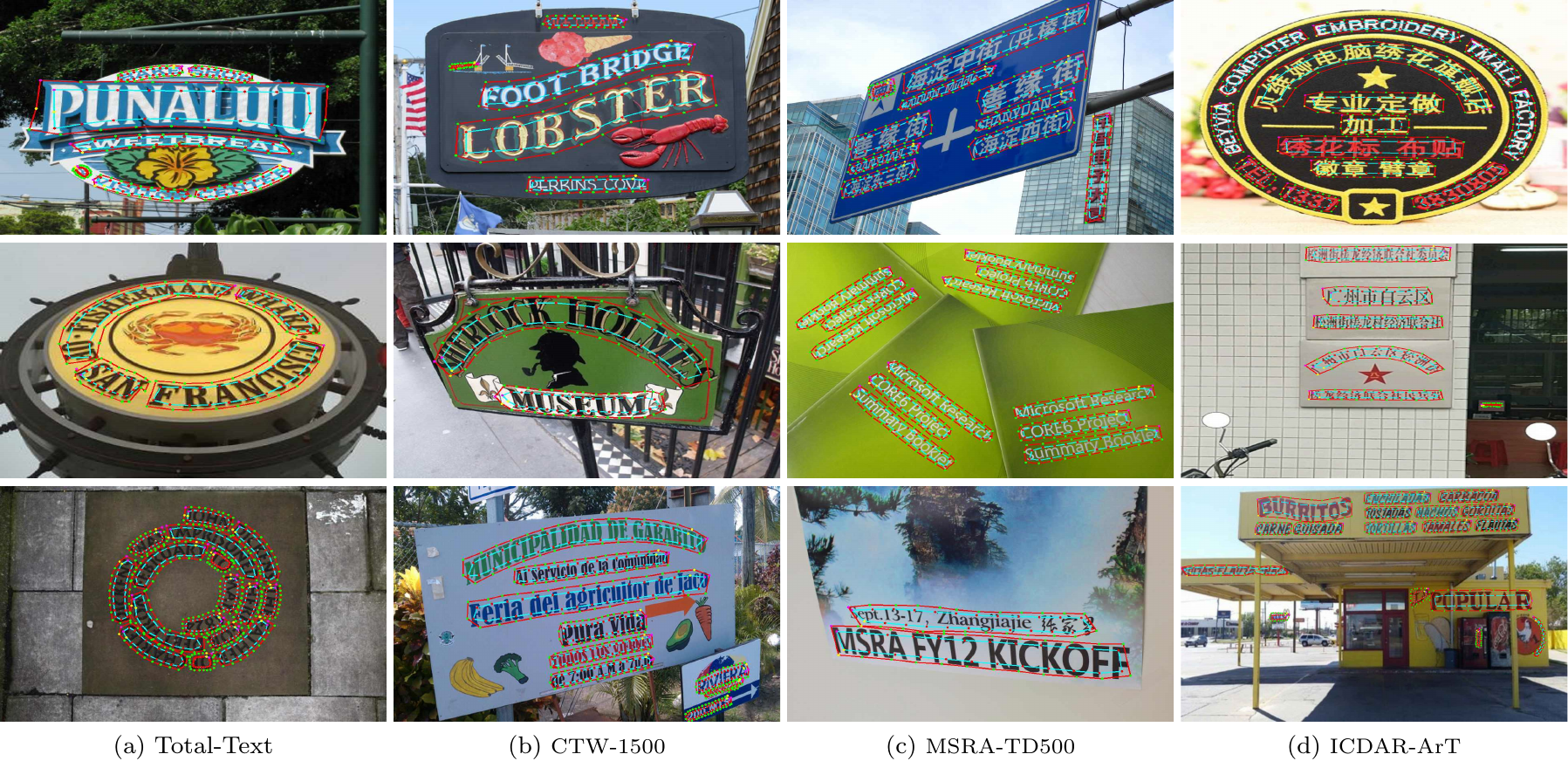}
	\centering
	\caption{ Visual experimental results. The blue contours are boundary proposals, and the red contours are the detection boundaries.}
	\label{fig:result_vis}
	\vspace{-1.2em}
\end{figure*}

\noindent\textbf{Influence of boundary energy loss (BEL).} We conduct ablation studies on Total-Text to verify the effectiveness of our BEL. As listed in Tab.~\ref{table:Ablation_BEL}, our method achieves the best performance by using both the direction field and boundary energy loss. BEL can enforce boundary optimization along the energy-decreasing direction in training. However, the direction of the energy decreasing is also the direction indicated by the direction field. Therefore, using BEL alone can also bring significant performance improvement (1.3\% in terms of F-measure), especially in terms of recall (2.22\%). In addition, BEL can improve the stability of training results. To illustrate this point, we counted the detection performance of the trained models from 300 epochs to 660 epochs and evaluated them every five epochs. So, we gain a total of 72 groups of data, and then we analyze the frequency distribution histogram of them, as shown in Fig.~\ref{fig:Frequency}. According to Fig.~\ref{fig:Frequency}, we can find the performance of the trained models is significantly improved as a whole with BEL. The experimental results in Tab.~\ref{table:iteration} also verify this point.

\begin{small}
	\begin{table}[tbp]
		\begin{center}
			\renewcommand{\arraystretch}{1.3}
			\caption{Experimental results on Total-Text for different resolution of feature map $ F $. “Res50” denotes using ResNet-50 as a backbone, and  “Res18” denotes using ResNet-18 as a backbone.}\label{table:rersolution}
			\begin{tabular}{c|cccc}
				\hline
				\textbf{Resolution}
				&\textbf{Recall}
				& \textbf{Precision} & \textbf{F-measures}&
				\textbf{FPS}\\
				\hline
				Res18 (1s)&83.68&89.99&86.72&16.0\\
				Res18 (2s)&83.27&89.04&86.06&25.2\\
				Res18 (4s)&79.55&90.38&84.62&32.5\\
				\hline
				\hline
				Res50 (1s)&85.29&89.86 &87.52&12.0\\
				Res50 (4s)&80.00&88.98&84.25&23.3\\
				\hline
			\end{tabular}
		\end{center}%
		\vspace{-1.8em}
	\end{table}
\end{small}

\noindent\textbf{Influence of different resolutions of the feature map.} We have conducted experiments to explore the influence of using the different resolution feature map $ F $, as listed in Tab.~\ref{table:rersolution}. In Tab.~\ref{table:rersolution}, “1s”, “2s”, and “4s” mean the width and height of output feature map $ F $ is 1/1, 1/2, and 1/4 of the testing image. From Tab.~\ref{table:rersolution}, we can see that 
the detection performance will decrease along with the resolution of output feature map $ F $ decreasing, but the detection speed is significantly improved. When the resolution of $ F $ decreases to “4s”, the detection performance decreases dramatically (2.1\% in terms of F-measure with Res18), especially the recall decreases by 4.13\%.
This is because the low resolution of $ F $ will cause some small text to be lost. Notably, the area of the connected regions for generating boundary proposals should be at least the number of control points in our method. However, our method with Res18 (4s) still achieves real-time detection efficiency (32.5 FPS) and impressive detection performance (84.62\% in terms of F-measure), which outperforms the majority of the state-of-the-art methods, \eg, DB~\cite{DB}, FCENet~\cite{FCENet}, TextRay~\cite{TextRay}.

\begin{small}
	\begin{table}[tbp]
		\begin{center}
			\renewcommand{\arraystretch}{1.2}
			\caption{Experimental results on Total-Text for different scale of input test image.}	\label{table:scale}
			\begin{tabular}{c|c|cccc}
				\hline
				\textbf{Resolution}& \textbf{Scale}
				&\textbf{Recall}
				& \textbf{Precision} & \textbf{F-measure}&
				\textbf{FPS}\\
				\hline
				Res18 (1s)&1024&83.68&89.99&86.72&16.0\\
				Res18 (1s)&800&82.72&88.12&85.34&20.5\\
				Res18 (2s)&1024&83.27&89.04&86.06&25.2\\
				Res18 (2s)&800&81.44&88.64&84.89&28.2\\
				Res18 (2s)&640&76.93&88.53&82.33&33.8\\
				Res18 (4s)&1024&79.55&90.38&84.62&32.5\\
				Res18 (4s)&800&77.50&90.22&83.38&35.3\\
				\hline
				\hline
				Res50 (1s)&1024&85.29&89.86 &87.52&12.0\\
				Res50 (1s)&800&81.68 &91.31 &86.23&16.1\\
				Res50 (4s)&1024&80.00&88.98&84.25&23.3\\
				Res50 (4s)&800&76.81&90.49&83.09&27.2\\
				
				\hline
			\end{tabular}
		\end{center}%
		\vspace{-1.8em}
	\end{table}
\end{small}

\begin{table*}[htbp]
	\begin{center}
		\renewcommand{\arraystretch}{1.0}
		\caption{Experimental results on Total-Text, CTW-1500 and MSRA-TD500. "Ext" means using the external dataset to pretrain the model.$ ^{\dagger} $ denotes the end-to-end scene text spotting. $^*$ denotes the method using ResNet50 with DCN~\cite{DCN} as a backbone. The best score is highlighted in \textbf{bold}.}	\label{table:tbSyn}
		\begin{tabular}{
				|m{2.95cm}<{\centering}
				|m{1.2cm}<{\centering}
				|m{0.6cm}<{\centering}
				||m{0.5cm}<{\centering}
				|m{0.5cm}<{\centering}
				|m{0.5cm}<{\centering}
				|m{0.5cm}<{\centering}
				||m{0.5cm}<{\centering}
				|m{0.5cm}<{\centering}
				|m{0.5cm}<{\centering}
				|m{0.5cm}<{\centering}
				||m{0.5cm}<{\centering}
				|m{0.5cm}<{\centering}
				|m{0.5cm}<{\centering}
				|m{0.5cm}<{\centering}|}
			\hline
			\multicolumn{1}{|c|}{ \multirow{2}*{ \textbf{Methods}}}
			&\multicolumn{1}{c|}{ \multirow{2}*{ \textbf{Published}}}
			&\multicolumn{1}{c||}{ \multirow{2}*{ \textbf{Ext}}}
			& \multicolumn{4}{c||}{\textbf{Total-Text}} 
			& \multicolumn{4}{c||}{\textbf{CTW-1500}}
			& \multicolumn{4}{c|}{\textbf{MSRA-TD500}}\\
			\cline{4-15}
			&&&\textbf{R}
			& \textbf{P} & \textbf{F}&
			\textbf{FPS}&
			\textbf{R}& 
			\textbf{P} & \textbf{F}&
			\textbf{FPS}&
			\textbf{R}& 
			\textbf{ P} & \textbf{F} &
			\textbf{FPS}\\
			\hline
			
			SegLink \cite{SegLink}&CVPR'17&Syn&- &- &-&-&- &- &-&- &70.0 &86.0 &77.0&8.9\\
			MCN \cite{MCN}& CVPR'18&Syn&- &- &- &-&- &- &-&- &79 &88 &83&-\\
			LSAE\cite{CVPR19_LSA}&CVPR'19&Syn &- &- &- &- &77.8 &82.7 & 80.1&- &81.7& 84.2 &82.9&-\\
			ATTR\cite{CVPR19_ATRR}&CVPR'19 &-&76.2 &80.9 & 78.5 &10.0 &- &- &- &- &82.1 &85.2 & 83.6 &-\\
			MSR\cite{MSR}&IJCAI'19 &Syn&73.0 &85.2 & 78.6 &- &79.0 &84.1 & 81.5 &- &76.7 &87.4 &81.7&-\\
			CSE\cite{CVPR19_CSE}&CVPR'19&MLT &79.7&81.4&80.2&0.4&76.1&78.7&77.4&0.38&- &- &-&-\\
			TextDragon$ ^{\dagger} $\cite{TextDragon}&ICCV'19&MLT$^{ +} $&75.7&85.6&80.3&-&82.8&84.5&83.6&-&- &- &-&-\\
			TextField\cite{TextField}&TIP'19&Syn &79.9&81.2&80.6&-&79.8&83.0&81.4&-&75.9 &87.4 & 81.3&5.2\\
			PSENet-1s \cite{CVPR19_PSENet}&CVPR'19&MLT &78.0&84.0&80.9&3.9&79.7&84.8& 82.2&3.9&- &- &-&-\\
			SegLink++ \cite{SegLink++}&PR'19&Syn&80.9&82.1& 81.5&-&79.8&82.8&81.3&-&- &- &-&-\\
			LOMO\cite{CVPR19_LOMO}&CVPR'19&MLT$ ^{+ }$ &79.3&87.6&83.3&-&76.5&85.7&80.8 &-&- &- &-&-\\
			CRAFT \cite{CRAFT}&CVPR'19&MLT &79.9&87.6&83.6&-&81.1&86.0&83.5&-&78.2 & 88.2 &82.9&8.6\\
			DB$ ^{*} $\cite{DB}&AAAI'20 &Syn&82.5 &87.1  &84.7 &{32.0}&80.2 &86.9 &83.4&{22.0}&79.2 &91.5 &84.9&32.0\\
			PAN\cite{PSENet_v2}&ICCV'19 &Syn&81.0 &89.3  &85.0&\textbf{39.6} &81.2 &86.4 &83.7&\textbf{39.8}&83.8 &84.4  &84.1&30.2\\
			TextPerception$ ^{\dagger} $ \cite{TextPerception}&AAAI'20&Syn &81.8&88.8&85.2&-&81.9&87.5&84.6&- &-  &- &-&-\\
			ContourNet~\cite{ContourNet}&CVPR'20&-&83.9&86.9&85.4&3.8&84.1&83.7&83.9 &4.5&-&-&-&-\\
			ABCNet$ ^{\dagger} $ \cite{ABCNet} &CVPR'20&MLT$ ^{+} $ &81.3 &87.9 &84.5 & 9.5&83.4 &84.4 &81.4 & 9.5&-&- &-  &-\\
			DRRG~\cite{DRRG}& CVPR'20&MLT&84.9&86.5&85.7&- &83.0&85.9&84.5&-&82.3&88.1&85.1&-\\
			Boundary$ ^{\dagger} $\cite{Boundary}&AAAI'20&Syn &85.0 &88.9  &87.0&- &- &- &- &-&- &-  &- &-\\
			TextRay \cite{TextRay}&MM'20&ArT$ ^{-} $&77.9 &83.5 &80.6  &-&80.4 &82.8 &81.6&-&-&-&-&-\\
			TextFuseNet \cite{TextFuseNet}&IJCAI'20&Syn &83.2 &87.5 &85.3  &7.1&\textbf{85.0} &85.8 &85.4&7.3&-&-&-&-\\
			Mask TextSpotter v3\cite{Mask_TextSpotter_v3}
			&ECCV'20&MLT&- &- &- &-&- &- &-&-&77.5 &90.7 &83.5 &-\\
			TextMountain\cite{TextMountain}&PR'21&MLT&- &- &- &-&82.9 &83.4 &83.2&-&- &- &- &-\\
			MOST\cite{MOST}&CVPR'21&Syn &- &- &-&- &-  &- &-&-&82.7&90.4&86.4&-\\
			PCR(Res50)\cite{PCR}&CVPR'21&- &80.2 &86.1 &83.1&-&79.8 &85.3 &82.4 &-&77.8 &87.6  &82.4 &-\\
			PCR(DLA34)\cite{PCR}&CVPR'21&MLT &82.0 &88.5 &85.2&-&82.3 &87.2 &84.7 &11.8&83.5 &90.8 &87.0&-\\
			FCENet \cite{FCENet}&CVPR'21&-&79.8&87.4&83.4&- &80.7 &85.7 &83.1&-&-&-&-&-\\
			CentripetalText \cite{CentripetalText}&NeurIPS'21&Syn&82.5 &90.5 &86.3&- &79.9 &88.3 &83.9 &-&82.5&90.0&86.1&34.8\\
			FCENet$ ^{*} $\cite{FCENet}&CVPR'21&-&82.5 &89.3 &85.8&- &83.4 &87.6 &85.5 &-&-&-&-&-\\
			Tang \etal \cite{FCBB}&CVPR'22&MLT&85.7 &90.7 &88.1 &12.9 &82.4  &88.1 &85.2 &12.9&-&-&-&-\\
			KPN(832)\cite{KPN}&TNNLS'22&MLT &85.6& 88.7 &87.1 &15.0
			&84.2 &84.3 &84.3 &16.3  &- &- &- &-\\
			EMA(DLA-34)\cite{EMA}&TIP'22&Syn &83.3& 88.2 &85.6 &24.2&82.1 &86.1  &84.1 &32.3  &81.1 &88.7.5 &84.7 &24.2\\
			DBNet++(800$ ^{*} $)\cite{DB++}&TPAMI'22&Syn &83.2& 88.9 &86.0 &28
			&82.8 &87.9  &85.3 &26  &83.3 &91.5 &87.2 &29\\
			Wang \etal\cite{FuzzyNet}&TIP'23&- &79.9& 88.7 &84.1 &24.3
			&82.5 &85.3 &83.9 &25.1  &81.6 &89.3  &85.3 &25.4\\
			TextPMs(watershed)\cite{TextPMs}&TPAMI'23&MLT 
			&87.7 &90.0 &88.8 &7.0  &83.8 &87.8 &85.8 &9.1 &\underline{86.9} &91.0 &88.9 &10.6\\
			\hline 
			\textbf{TextBPN}\cite{TextBPN}&ICCV'21&Syn &84.65 &90.27 &87.37 &10.3 &81.45 &87.81 &84.51 &12.2 &80.68 &85.40 &82.97 &12.7\\
			\textbf{TextBPN}\cite{TextBPN}&ICCV'21&MLT &85.19 &90.67 &87.85 &10.7&83.60&86.45&85.00 &12.2&84.54&86.62&85.57&12.3\\
			\hline
			\hline
			\vspace{2pt}
			\textbf{Ours}(Res18-4s-1024)&-&MLT &81.90 &89.88 &85.70 &\underline{32.5} &81.62 &87.55&84.48 &\underline{35.3} &\textbf{87.46}&92.38&89.85&\textbf{38.5}\\
			\textbf{Ours}(Res50-1s-1024)&-&-&85.29 &89.86 &87.52&12.0&81.12&88.08 &84.46&14.7&81.27&88.25&84.62&15.7\\
			\textbf{Ours}(Res50-1s-1024)&-&MLT&85.34 &91.81 &88.46&13.3&83.77 &87.30&85.50&14.1&85.40&89.23&87.27&15.2\\
			\textbf{Ours}(Res50-1s-1024$ ^{*} $)&-&MLT&\textbf{87.93} &\textbf{92.44} &\textbf{90.13}&13.2&\underline{84.71} &\textbf{88.34}&\textbf{86.49}&16.5&86.77&\textbf{93.69}&\textbf{90.10}&15.3\\
			\hline
		\end{tabular}
	\end{center}%
	\vspace{-1.8em}
\end{table*}

\noindent\textbf{Influence of different scales of testing images.} We have conducted experiments to explore the influence of different scales for testing images. We limit testing images to different scales and feed them to the network for processing. The detailed experimental results are listed in Tab.~\ref{table:scale}. From Tab.~\ref{table:scale}, we can find that the performance will degrade along with the decrease of the resolution of the testing image, while the detection speed is enhanced significantly. 
Compared with the scale 1024, using the scale 800 will bring about 1\% to 1.5\% loss of detection performance in terms of F-measure, and improve the detection speed by about 4 to 5 FPS. When the scale of the testing image decreases to 800 with Res18 (4s), the detection speed of our method achieves real-time speed (35.3 FPS) while the performance (83.38\% in terms of F-measure) is still competitive against other state-of-the-art methods~\cite{FCENet, PCR}. Note that the performance of Res50 (4s) is slightly lower than Res18 (4s), which may be because we use a larger batch size to train Res18 (4s) on a single GPU.

\subsection{Comparisons with State-of-the-art Methods}\label{exp_Comparisons}
We compare our proposed method with other methods
on five standard benchmarks, including three benchmarks for curved text (Total-Text, CTW-1500, and ICDAR-ArT) and two multi-language and multi-oriented benchmarks for long text lines (MSRA-TD500  and ICDAR-MLT17).

\noindent\textbf{Total-Text}. During testing, we set both sides of the image in the range of (640, 1024) while keeping its aspect ratio. The threshold $ th_d $ and $ th_s $ are set to 0.325 and 0.85, respectively. The quantitative results are listed in Tab.~\ref{table:tbSyn}. From Tab.~\ref{table:tbSyn}, we can find that our method (``Res50-1s-1024'') achieves $87.52\%$ in terms of F-measure without any pre-training, and $88.46\%$ in terms of F-measure when pre-trained on MLT17. In addition, with a lightweight backbone (``Res18-4s"), our method achieves a real-time detection speed (32.5 FPS) and comparable performance against PCR (DLA34)~\cite{PCR} (85.70\% vs. 85.2\% F-measure) and FCENet$ ^{*} $~\cite{FCENet} (85.70\% vs. 85.8\% F-measure).
When using ResNet-50 with DCN~\cite{DCN} as a backbone like DB$ ^{*} $~\cite{DB}  and FCENet$ ^{*} $~\cite{FCENet}, our method achieves  top performance (90.13\% F-measure) and outperforms all the state-of-the-art methods with a great margin. For example, our method (``Res50-1s-1024$ ^{*} $") outperforms the FCENet$ ^{*} $~\cite{FCENet} by 4.33\% in terms of F-measure, and outperforms the DB$ ^{*} $~\cite{DB} by 5.43\% in terms of F-measure. Qualitative visible results are shown in Fig.$ \, $\ref{fig:result_vis} (a) and (b).

\begin{figure*}[htbp]
	\subfigcapskip=-3pt
	\centering
	\includegraphics[width=1.0\linewidth]{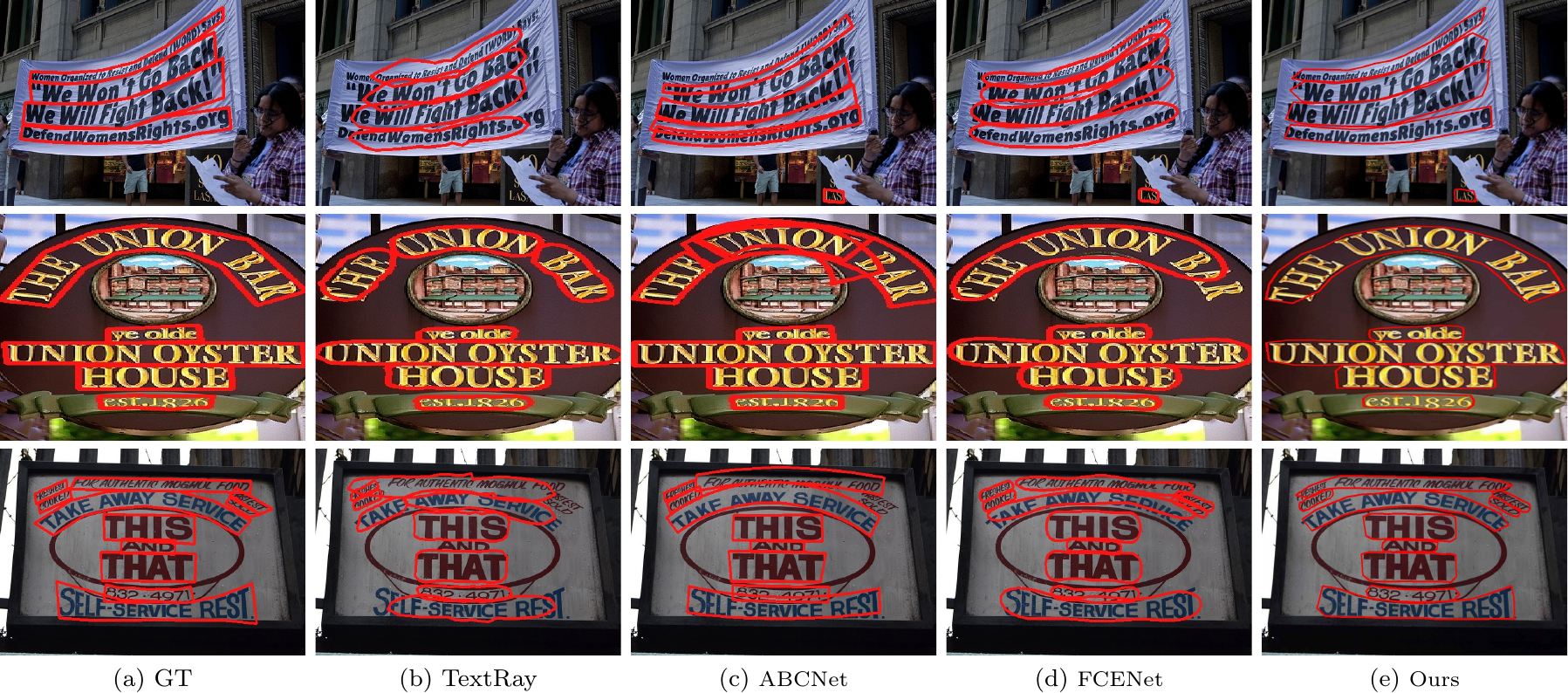}
	\centering
	\caption{Qualitative comparisons with TextRay~\cite{TextRay}, ABCNet~\cite{ABCNet}, and FCENet~\cite{FCENet} on selected challenging samples from CTW-1500. The images (a)-(d) are borrowed from FCENet~\cite{FCENet}.}\label{fig_ch}
	\vspace{-1.8em}
\end{figure*}

\noindent\textbf{CTW-1500}. During testing, we set both sides of the image in the range of (640, 1024) while keeping its aspect ratio. The threshold $ th_d $ and $ th_s $ are set to 0.375 and 0.8, respectively. Representative visible results are shown in Fig.~\ref{fig:result_vis} (c) and (d), which indicate our method precisely detects boundaries of the long curved text. The quantitative results are listed in Tab.$ \ $\ref{table:tbSyn}. Compared with the state-of-the-art methods~\cite{DB,PSENet_v2,ContourNet}, our approach (``Res50-1s-1024$ ^{*} $") achieves promising performance of both Recall ($84.71\%$) and F-measure ($ 86.49\% $). Specifically, our method greatly outperforms TextField~\cite{TextField}  and DB~\cite{DB} by $5.09\%$ and $3.09\%$ in terms of F-measure, respectively. When equipped with a  lightweight backbone (``Res18-4s"), our method achieves a competitive detection speed and improved performance (84.48\% F-measure, 34.8 FPS) against DB~\cite{DB} (83.4\% F-measure, 22.0 FPS) and PAN~\cite{PSENet_v2} (83.7\% F-measure, 39.8 FPS). Notably, the detection efficiency has been slightly improved when ResNet-50 with DCN is used as the backbone. It's because the quality of prior information (\ie, distance field, classification map) is improved, resulting in less noise and false detection, which will improve the detection speed.

\begin{table}[tbp]
	\begin{center}
		\renewcommand{\arraystretch}{1.3}
		\caption{Comparisons with related works on ICDAR-ArT. Some experimental result are from \cite{PCR}}
		\label{table:ArT}
		\begin{tabular}{c|c|c|ccc}
			\hline
			\textbf{Methods}&\textbf{Published}&Ext& \textbf{R}& \textbf{P} & \textbf{F}\\
			\hline
			TextRay(Res50) \cite{TextRay}&MM'20&$ \checkmark $ &58.60 &75.97 &66.17\\ 
			PCR(DLA-34) \cite{PCR} &CVPR'21&$ \times $&65.0 &83.6 &73.1\\
			PCR(DLA-34) \cite{PCR} &CVPR'21&$ \checkmark $&66.1 &84.0 &74.0\\
			EMA(DLA-34)\cite{EMA}&TIP'22&$ \checkmark $&68.7&80.8&74.3\\
			Wang \etal\cite{FuzzyNet}&TIP'23&$ \times $ &60.6 &78.5  &68.4\\
			\hline
			\hline
			\textbf{Ours}(Res50-1s) &-&$ \checkmark $&71.07&81.14 &75.77\\
			\textbf{Ours}(Res50-1s$ ^{*} $) &-&$ \checkmark $&\textbf{77.05}&\textbf{84.48}&\textbf{80.59}\\
			\hline
		\end{tabular}
	\end{center}%
	\vspace{-2.0em}
\end{table}

\begin{table}[tbp]
	\begin{center}
		\renewcommand{\arraystretch}{1.3}
		\caption{Experimental results on ICDAR-MLT17. $^*$ denotes the method using ResNet50 with DCN~\cite{DCN} as a backbone. Here, the "Ext" means using the SynthText for pre-training}
		\label{table:MLT}
		\begin{tabular}{c|c|c|ccc}
			\hline
			\textbf{Methods}&\textbf{Published}&Ext&\textbf{R}& \textbf{P} & \textbf{F}\\
			\hline
			Ma et al. \cite{RRPN} &TMM'18&$  \checkmark $&55.50 &71.17 &62.37\\ 
			He et al. \cite{MOML} &TIP'18&$  \checkmark $&57.9 &76.7  &66.0\\ 
			Border \cite{ASTD} &ECCV'18&$  \checkmark $&60.6 &73.9  &66.6\\ 
			Corner.\cite{corner} &CVPR'18&$  \checkmark $&55.6 &\textbf{83.8} &66.8\\
			FOTS\cite{FOTS} &CVPR'18&$  \checkmark $&57.51 &80.95 &67.25\\
			DRRG~\cite{DRRG}&CVPR'20&$  \checkmark $&61.04&74.99&67.31\\
			LOMO\cite{CVPR19_LOMO} &CVPR'19&$  \checkmark $&60.6 &78.8 &68.5\\
			SAST~\cite{SAST} &MM'19&$  \checkmark $&67.56 &70.00 &68.76\\
			SPCNet \cite{SPCNet} &AAAI'19&$  \checkmark $&66.9 &73.4  &70.0\\ 
			PSENet-1s \cite{CVPR19_PSENet}&CVPR'19&$  \checkmark $&68.21&73.77&70.88\\
			DB-Res18$ ^{*} $\cite{DB}&AAAI'20&$  \checkmark $&63.8&81.9 &71.7\\
			DB-Res50$ ^{*} $\cite{DB}&AAAI'20&$  \checkmark $&67.9&83.1 &74.7\\
			\hline
			\hline
			\textbf{Ours}(Res50-1s) &-&$  \times $&65.67 &80.49 &72.33\\
			\textbf{Ours}(Res50-1s$ ^{*} $) &-&$  \checkmark $&\textbf{72.10} &83.74 &\textbf{77.48}\\
			\hline
		\end{tabular}
	\end{center}%
	\vspace{-1.8em}
\end{table}

\noindent\textbf{MSRA-TD500}. Due to a small amount of training data, we train the model on MSRA-TD500 for 1200 epochs. During testing, we set both sides of the image in the range of (640, 960) while keeping its aspect ratio. The threshold $ th_d $ and $ th_s $ are set to 0.35 and 0.9, respectively. The quantitative comparisons are listed in Tab.$ \, $\ref{table:tbSyn}. According to Tab.~\ref{table:tbSyn}, our method
successfully detects long text lines of arbitrary orientations and sizes.  
When equipped with a  lightweight backbone (``Res18-4s"), our method achieves $89.85\%$ in terms of F-measure and 38.5 FPS, which outperforms other state-of-the-art methods, such as DB~\cite{DB} (84.9\% F-measure, 32 FPS), PAN~\cite{PSENet_v2} (84.1\% F-measure, 30.2 FPS), \etc When using ResNet-50 with DCN~\cite{DCN} as a backbone, our method (``Res50-1s-1024$ ^{*} $") achieves the promising performance (93.69\% Precision, 90.10\% F-measure).

\noindent\textbf{ICDAR-ArT}. To demonstrate the generalization ability of the proposed method, we test our model on the ICDAR-ArT dataset, containing a lot of multi-lingual curved text instances from the complex scene. During testing, we set both sides of the image in the range of (960, 2880) while keeping its aspect ratio. The threshold $ th_d $ and $ th_s $ are set to 0.4 and 0.8, respectively. As listed in Tab.~\ref{table:ArT}, our method can
boost the F-measure from 74.0\% to 80.59\%, compared with
the recent state-of-the-art method PCR~\cite{PCR}. Specifically, our method (``Res50-1s$ ^{*} $") outperforms PCR~\cite{PCR}  by 6.59\% in terms of F-measure and by $10.95\%$ in terms of Recall, respectively. Fig.~\ref{fig:result_vis} (c) shows the qualitative results. Fig.~\ref{fig_ch} (b) and (e) compare the detection results of TextRay~\cite{TextRay} and our method, respectively.

\begin{figure}[tbp]
	\subfigcapskip=-3pt
	\centering
	\includegraphics[width=1.0\linewidth]{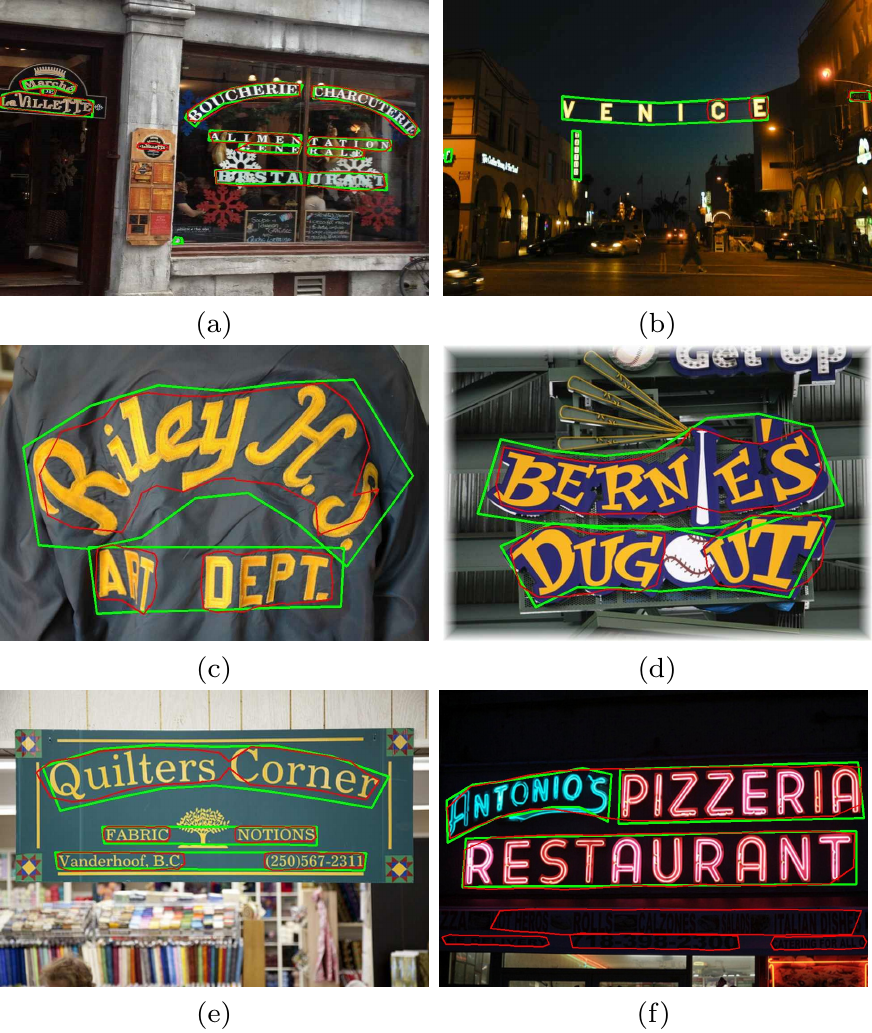}
	\centering
	\caption{ Some visual results of failure cases. Green contours: ground truths; red contours: detections of our methods. Fig. (a-b) are selected from Total-Text, and Fig. (c-f) are selected from CTW-1500.}
	\label{fig:weakness}
	\vspace{-1.8em}
\end{figure}

\noindent\textbf{ICDAR-MLT17}. During testing, we set both sides of the image in the range of (960, 2048) while keeping its aspect ratio. The threshold $ th_d $ and $ th_s $ are set to 0.5 and 0.85, respectively. As listed in Tab.~\ref{table:MLT}, our method achieves impressive performance (77.48\% in the terms of F-measure). Specifically, our method (``Res50-1s$ ^{*} $") outperforms DB-Res50$ ^{*} $~\cite{PCR} by $2.78\%$ in terms of F-measure  and  by $4.2\%$ in terms of Recall , respectively. Our method also significantly outperforms other state-of-the-art methods, such as LOMO~\cite{CVPR19_LOMO}, SAST~\cite{SAST}, SPCNet~\cite{SPCNet}, PSENet~\cite{CVPR19_PSENet}.

\noindent\textbf{Visual Comparison.}
As shown in Fig.~\ref{fig_ch}, we give a more intuitive comparison between our method and the state-of-the-art methods, including TextRay~\cite{TextRay}, ABCNet~\cite{ABCNet}, and FCENet~\cite{FCENet}.  Compared with other contour-based methods (\ie, TextRay~\cite{TextRay}, ABCNet~\cite{ABCNet}, and FCENet~\cite{FCENet}, our method can generate more accurate text boundaries for explicitly modeling irregular text instances. Fig.~\ref{fig_ch} shows the effectiveness of our method for highly-curved text detection.

\subsection{Weakness}
Some failure cases are shown in Fig.~\ref{fig:weakness}. 
Although our method achieves superior performance, there is still a chance of failure cases due to the high complexity of scene images, such as object occlusion in Fig.~\ref{fig:weakness} (a) (d) and (e), large character spacing in Fig.~\ref{fig:weakness} (a) (b) and (c). These cases are still very challenging and nontrivial in the paradigm of text detection. Moreover, there are some “false
detections” caused by unreasonable or missing annotations, as shown in Fig.~\ref{fig:weakness} (e) and (f).


\section{Conclusion} \label{Conclusion}
In this paper, we present a comprehensive coarse-to-fine framework based on boundary learning for arbitrary shape text detection. Our approach can efficiently and accurately locate text boundaries without complex  post-processing. The proposed framework comprises a feature extraction backbone, a boundary proposal module, and an iteratively optimized boundary transformer module. Our method explicitly models the text boundary using an innovative iterative boundary transformer in a coarse-to-fine manner, which enables us to directly obtain accurate text boundaries and eliminate the need for complex post-processing. Our experimental results on publicly available datasets demonstrate the state-of-the-art performance and efficiency of our method. In the future, we plan to develop a text spotting framework based on our current work for arbitrary shape text.

\bibliographystyle{IEEEtran}
\bibliography{IEEEbib}

 \end{document}